\documentclass[10pt,twocolumn,letterpaper]{article}

\usepackage[pagenumbers]{cvpr} 

\definecolor{cvprblue}{rgb}{0.21,0.49,0.74}
\usepackage[pagebackref,breaklinks,colorlinks,allcolors=cvprblue]{hyperref}

\usepackage{makecell} 
\usepackage{algorithm}      
\usepackage{algpseudocode}  

\usepackage{graphicx}

\usepackage{hyperref}
\usepackage{url}
\usepackage{booktabs} 
\usepackage{wrapfig} 
\newcommand{\tablestyle}[2]{\setlength{\tabcolsep}{#1}\renewcommand{\arraystretch}{#2}\centering\footnotesize}

\usepackage{colortbl} 
\usepackage{xcolor}

\usepackage{caption}
\usepackage{subcaption}

\AtEndPreamble{
    \usepackage[capitalize]{cleveref}
    \crefname{section}{Sec.}{Secs.}
    \Crefname{section}{Section}{Sections}
    \crefname{table}{Tab.}{Tabs.}
    \Crefname{table}{Table}{Tables}
}

\usepackage{hyperref}
\usepackage{url}

\usepackage{graphicx}
\usepackage{hyperref}
\usepackage[capitalize,noabbrev]{cleveref}

\usepackage{caption}
\usepackage{subcaption}

\usepackage{url}
\usepackage{booktabs} 
\usepackage{wrapfig} 

\usepackage{booktabs}       

\usepackage{color,xcolor}
\usepackage{epsfig}
\usepackage{graphicx}

\usepackage{adjustbox}
\usepackage{array}
\usepackage{booktabs}
\usepackage{colortbl}
\usepackage{hhline}
\usepackage{multirow}
\usepackage{tabularx}
\usepackage{makecell}
\usepackage{float}
\usepackage{subcaption}  

\usepackage{bm}
\usepackage{nicefrac}
\usepackage{microtype}

\usepackage{changepage}
\usepackage{extramarks}
\usepackage{fancyhdr}
\usepackage{soul}
\usepackage{xspace}

\usepackage{url}

\usepackage{enumerate}
\usepackage{enumitem}

\usepackage{etoolbox,siunitx}
\usepackage{overpic}
\usepackage{bbding}

\def\module{MagicView}
\def\modulespace{MagicView }
\def\x{{\mathbf{x}}}

\def\paperID{17672} 
\def\confName{CVPR}
\def\confYear{2026}

\title{\module: Multi-View Consistent Identity Customization\\ via Priors-Guided In-Context Learning}

\author{%
Hengjia Li$^{1,*}$, Jianjin Xu$^{2,*}$, Keli Cheng$^{3}$, Lei Wang$^{3}$, \\Ning Bi$^{3}$, Boxi Wu$^{1, \dag}$, Deng Cai$^{1}$, Fernando De la Torre$^{2}$ 
 \\\\
 $^1$Zhejiang University\qquad$^2$Carnegie Mellon University\qquad
 $^3$Qualcomm Inc.
 \\ 
 {\tt\small lihengjia98@gmail.com}\\ 
}

\begin{document}

\twocolumn[{
\renewcommand\twocolumn[1][]{#1}
\maketitle
\begin{center}
    \centering
    \vspace*{-.8cm}
    \includegraphics[width=\textwidth]{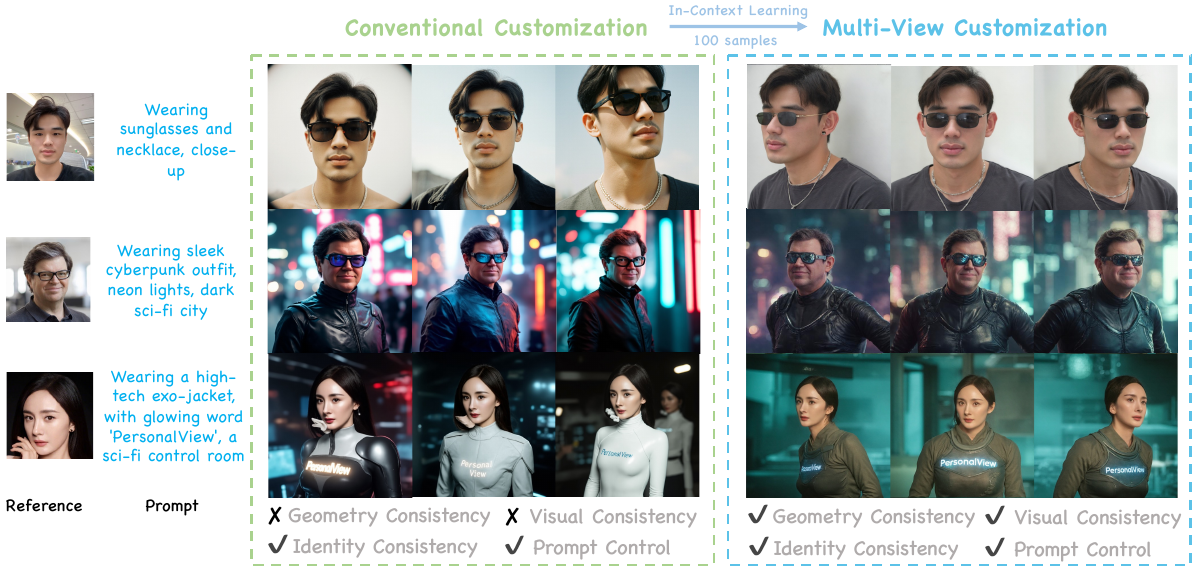}
    \vspace*{-.6cm}
        \captionof{figure}{\textbf{\modulespace compared to conventional customization methods.}
  \modulespace generates personalized images consistent with multiple views given one reference image.
  Conventional methods like PULID~\citep{guo2024pulid} have limited control over the viewpoint in the prompt (\textit{i.e.}, left, middle, and right view) and do not have multi-view consistency.}
\label{fig:teaser}
\end{center} 
}]

\begin{abstract}

Recent advances in personalized generative models have demonstrated impressive capabilities in producing identity-consistent images of the same individual across diverse scenes. However, most existing methods lack explicit viewpoint control and fail to ensure multi-view consistency of generated identities.
To address this limitation, we present \textbf{\module}, a lightweight adaptation framework that equips existing generative models with multi-view generation capability through 3D priors–guided in-context learning.
While prior studies have shown that in-context learning preserves identity consistency across grid samples, its effectiveness in multi-view settings remains unexplored. Building upon this insight, we conduct an in-depth analysis of the multi-view in-context learning ability, and design a conditioning architecture that leverages 3D priors to activate this capability for multi-view consistent identity customization.
On the other hand, acquiring robust multi-view capability typically requires large-scale multi-dimensional datasets, which makes incorporating multi-view contextual learning under limited data regimes prone to textual controllability degradation. To address this issue, we introduce a novel Semantic Correspondence Alignment loss, which effectively preserves semantic alignment while maintaining multi-view consistency.
Extensive experiments demonstrate that \modulespace substantially outperforms recent baselines in multi-view consistency, text alignment, identity similarity, and visual quality, achieving strong results with only 100 multi-view training samples.


\end{abstract}

\section{Introduction}
\begin{table*}[t]
    \centering
    \caption{\textbf{Comparison with previous methods.} In comparison, our method is capable of generating geometrically and visually consistent multi-view images while preserving human identity and prompt controllability, all without requiring large-scale multi-view datasets or test-time training.}
    \tablestyle{2pt}{1}
    \begin{tabular}{ccccccc}
\toprule
Method & \makecell[c]{Geometry \\ Consistency} & \makecell[c]{Visual \\Consistency}  & \makecell[c]{w/o Large \\MV Dataset} & \makecell[c]{w/o Test-Time \\Training} & \makecell[c]{Human Identity \\Consistency} & \makecell[c]{Prompt \\Control}\\
\midrule
\makecell[c]{Conventional Customization\\ \scriptsize(\textit{e.g.}, PuLID \citep{guo2024pulid})} &\XSolidBrush & \XSolidBrush&\Checkmark & \Checkmark & \Checkmark & \Checkmark   \\
\midrule
\makecell[c]{Image to Multi-View \\\scriptsize(\textit{e.g.}, Era3D \citep{li2024era3d})} &\Checkmark & \XSolidBrush& \XSolidBrush & \Checkmark &\Checkmark & \XSolidBrush   \\
\midrule
\makecell[c]{Unified Multi-Modal Model \\\scriptsize(\textit{e.g.}, BAGEL \citep{deng2025emerging})} &\XSolidBrush & \XSolidBrush& \Checkmark & \Checkmark &\Checkmark & \Checkmark   \\
\midrule
\makecell[c]{CustomDiffusion360 \\ \scriptsize\citep{kumari2024customizing}}&\XSolidBrush&\Checkmark & \Checkmark&\XSolidBrush&\XSolidBrush&\Checkmark \\
\midrule
\makecell[c]{ICLoRA \\ \scriptsize\citep{lhhuang2024iclora}}&\XSolidBrush&\Checkmark & \XSolidBrush&\XSolidBrush&\Checkmark&\Checkmark \\
\midrule
\modulespace &\Checkmark &\Checkmark&\Checkmark & \Checkmark&\Checkmark&\Checkmark
 \\ 
\bottomrule
\end{tabular}
\centering
\vspace{-0.2cm}
\label{tab:sum}
\end{table*}


Recent years have seen remarkable progress in human image customization~\citep{li2024photomaker, guo2024pulid, wang2024instantid, ipadapter, li2024personalvideo, li2025magicid, yang2024lora}, enabling the generation of personalized images from user-provided photographs. Given a single portrait, these methods can synthesize customized depictions of the same individual in diverse contexts, such as sitting on a beach or smiling in a meadow.
However, a key challenge remains — achieving multi-view consistent customization that preserves identity and appearance across varying viewpoints.

When users wish to change the view of a customized human while synthesizing it in a novel context (\cref{fig:teaser}), viewpoint-specific prompts such as “from the left view” offer limited control over diverse perspectives. The generated results often lack inter-view coherence, exhibiting inconsistencies in geometry and visual appearance across body shape, facial features, expressions, clothing, and background. This raises a key question: how can multi-view consistency be preserved during personalized image generation?

In this paper, we introduce a new task, Multi-View Customization (MVC), generating multi-view human images conditioned on a single photo. Without test-time tuning, MVC produces identity-faithful results with strong geometry and visual consistency across views, while maintaining flexible prompt-based control (\cref{tab:sum}). Beyond personalized creation, this paradigm can be extended to broader applications such as 3D modeling and reconstruction.


The core of this task is how to control the generation of multiple views while maintaining geometric and visual consistency. We are inspired by the in-context learning ability of the DiT-based models, such as FLUX~\citep{flux1-dev}. Prior works~\citep{lhhuang2024iclora, kang2025flux} have demonstrated the ability to generate a grid of images with roughly consistent visual content. However, they unexplore the potential to preserve the consistency in multi-view scenarios. It motivates us to investigate whether the model has a strong multi-view capability to foster more explicit multi-view consistency.

To this end, we employ in-context depth maps as 3D priors to activate FLUX’s capability for geometrically and visually consistent generation. Specifically, we propose a multi-view depth–based framework, termed \module.
Our goal is to generate multi-view depth maps consistent with the given prompts, serving as contextual priors for subsequent synthesis. In the first stage, we leverage a pre-trained customization model~\citep{guo2024pulid} for preliminary sampling and use the SMPL model~\citep{goel2023humans} to fit and render multi-view depth maps. These depth maps are then arranged into a four-panel grid, providing in-context conditioning for a depth-guided model to synthesize multi-view consistent customized images.

However, acquiring robust multi-view customization capability typically demands large-scale multi-dimensional datasets, making multi-view contextual learning under limited data prone to textual controllability degradation.
To address this issue, we propose a Semantic Correspondence Alignment (SCA) loss tailored for DiT-based models to preserve the model’s intrinsic semantic controllability.
Specifically, SCA incorporates a frozen reference branch of the original model during training and enforces token-level correspondence alignment between the trainable and frozen branches across both text and visual modalities.
This design effectively maintains the model’s semantic precision while preserving its in-context learning capability for multi-view reasoning.

Our proposed \modulespace consistently preserves multi-view human identity requiring only a few training steps without large-scale multi-view human dataset.
It exhibits robust multi-view consistency across various attributes, including facial features, body gestures, background, and clothing, while concurrently preserving excellent prompt controllability and identity fidelity.
Extensive experimentation demonstrates its superior performance compared to existing novel view synthesis methods that typically necessitate extensive training on large-scale multi-view datasets.

In summary, our contributions are as follows.
\begin{itemize}
\item We introduce a novel task, Multi-View Customization conditioned on a single user-provided photograph without test-time tuning, which aims to generate multi-view consistent customized images with precise identity and diverse prompt-control.
\item We propose \module, leveraging in-context multiview priors to activate the in-context learning capabilities of DiT-based models with only a few training samples, facilitating the generation of geometry and visual consistent human images.
\item We introduce a Semantic Correspondence Alignment loss tailored for DiT-based models, which preserves the original model's semantic control capabilities with limited multi-view data.
\end{itemize}

\section{Related Works}
\subsection{Image Customization}
In the domain of Text-to-Image (T2I) generation, a variety of approaches have emerged to address identity (ID) customization~\citep{gal2022image, li2024photomaker, gal2024lcm, valevski2023face0, xiao2024fastcomposer, ma2024subject, peng2024portraitbooth, li2023few, li2024unihda, chen2023photoverse, jin2025scar,xu2025scalar,lan2025flux}. A seminal method in this line of work is Textual Inversion~\citep{gal2022image}, which encodes user-specific identity information into a dedicated token embedding while keeping the T2I model parameters fixed. To improve identity fidelity, 
In contrast, encoder-based paradigms aim to directly inject identity representations into the generation pipeline.
For instance, PhotoMaker~\citep{li2024photomaker} leverages large-scale identity datasets to construct robust ID embeddings from diverse image samples.
Similarly, PuLID~\citep{guo2024pulid} introduces a more precise ID supervision mechanism by minimizing identity loss between synthesized outputs and reference images.
Recently, CustomDiffusion360 enables explicit control over object viewpoints in the customization of text-to-image diffusion models; however, it requires test-time training and is limited to object-level customization rather than human subjects.

\subsection{Multi-View Images Synthesis}
Cross-view consistency plays a pivotal role in multi-view generation. MVFusion~\citep{tang2023MVDiffusion} initiates this direction by parallel multi-view image generation with correspondence-aware attention, which facilitates cross-view information exchange and supports textured scene mesh reconstruction. Building upon it, subsequent works~\citep{tseng2023consistent, kant2024spad, gu2024diffportrait3d, li2024gca, shen2024imagpose} incorporate epipolar constraints into diffusion models to enhance inter-view feature alignment. Zero123++~\citep{shi2023zero123++} adopts a tiled view representation, enabling single-pass generation over multiple views, a strategy later adopted in Direct2.5 and Instant3D for efficient view synthesis. 
Similarly, MVDream~\citep{shi2023mvdream} and Wonder3D~\citep{long2023wonder3d} leverage dedicated multi-view self-attention mechanisms to promote cross-view coherence. 
Meanwhile, other approaches~\citep{chen2024v3d, voleti2024sv3d, yu2024viewcrafter} exploit spatio-temporal priors from video diffusion models to ensure view consistency across frames.

\subsection{In-Context Learning}
Recent advancements in Text-to-Image (T2I) generation~\citep{openai2023dalle3, podell2023sdxl, esser2024scaling, black2024flux} have enabled the synthesis of identity-preserving subject views within a single $M{\times}N$ grid-structured mosaic, where each subview is prompted via carefully designed textual inputs. 
For example, by formulating the task as a grid-based image completion problem and replicating the subject image(s) within a structured mosaic layout,~\citep{kang2025flux} demonstrates strong identity-preserving capabilities without the need for additional training data, fine-tuning, or modifications during inference.
IC-LoRA~\citep{lhhuang2024iclora} explores in-context learning by fine-tuning a LoRA~\citep{hu2021lora} model on concatenated grid-based image-prompt pairs. However, they under-explore the multi-view scenarios which suffers from reduced visual consistency in transferring identity across views.


\section{Method}

\begin{figure*}[t]
    \centering    \includegraphics[width=\linewidth]{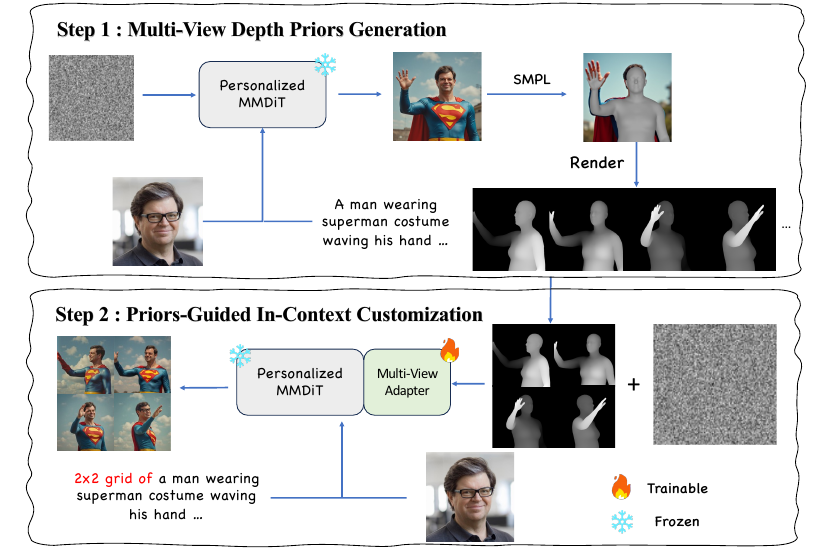}
    \vspace{-0.2cm}
    \caption{\textbf{Overview of \module.} In step 1, we use SMPL~\citep{goel2023humans} to fit the body mesh corresponding to the sample from the personalized generator~\citep{guo2024pulid}. Then we render the body mesh for multi-view depth maps. With the in-context depth priors, we can generate the multi-view customization images in step 2 using the personalized model with our control Adapter.}
    \label{fig:method}
    \vspace{-0.4cm}
\end{figure*}

\subsection{Preliminary}
\textbf{Diffusion Models.} Diffusion  models~\citep{sohl2015deep,ho2020denoising} are a class of likelihood-based generative models that produce data samples via a sequential denoising procedure originating from Gaussian white noise. During training, a predefined forward diffusion process is employed to transform clean observations $\mathbf{x}_0$ into a latent noise representation $\mathbf{x}_T \sim \mathcal{N}(\mathbf{0}, \mathbf{I})$ through the iterative injection of Gaussian perturbations across $T$ steps, forming a Markov chain structure, i.e., $\x_t = \sqrt{\alpha_t}\x_{0} + \sqrt{1 - \alpha_t}\epsilon$. The model is trained to learn the backward process, i.e., 

\begin{equation}
    \begin{aligned}
    p_\theta(\x_0 | \mathbf{c}) %
    = \int \Bigr [ p_{\theta} (\x_{T}) \prod p_{\theta}^t(\x_{t-1} | \x_t, \mathbf{c}) \Bigr ] d\x_{1:T}, 
    \end{aligned}\label{eq:diffusionformulation}
\end{equation}
Typically, the training objective maximizes the variational lower bound, which can be simplified to a simple reconstruction loss with the conditioning signal $\mathbf{c}$: 
\begin{equation}
    \begin{aligned}
     \mathcal{L}_{\text{diff}} = \mathbb{E}_{\x_t,t,\mathbf{c}, \epsilon \sim \mathcal{N} (\mathbf{0}, \mathbf{I})} [w_t||\epsilon - \epsilon_{\theta} (\x_t, t, \mathbf{c}) ||].
    \end{aligned}\label{eq:diffusion}
\end{equation}

\noindent
\textbf{Diffusion Transformers.}
A growing body of research has begun to adopt transformer~\citep{dit} architectures within text-to-image generative frameworks. Notably, models such as FLUX~\citep{flux1-dev} exemplify this trend by employing the MultiModal-Diffusion Transformer (MM-DiT) architecture. This design facilitates joint cross-modal interaction by performing attention over concatenated text and image embeddings, thereby enabling more effective integration of multimodal information during the generation process.

\begin{equation}    
    Q = [Q_{T};Q_{I}], K = [K_{T}; K_{I}], V = [V_{T}; V_{I}],
    \label{eq:attn_qkv}
\end{equation}
\begin{equation}
    \text{A}(Q, K, V) = W(Q,K)V = \text{softmax}\left(\frac{QK^T}{\sqrt{d}}\right)V,
    \label{eq:attn_eq}
\end{equation}
where $[;]$ is the concatenation, $Q$, $K$, and $V$ represent the key components of attention$-$query, key, and value, respectively; $Q_t$ and $Q_i$ correspond to the text and image query tokens; $W$ is the attention weight, and $A$ is the output correspondence.

\begin{figure*}[t]
    \centering
    \includegraphics[width=.9\linewidth]{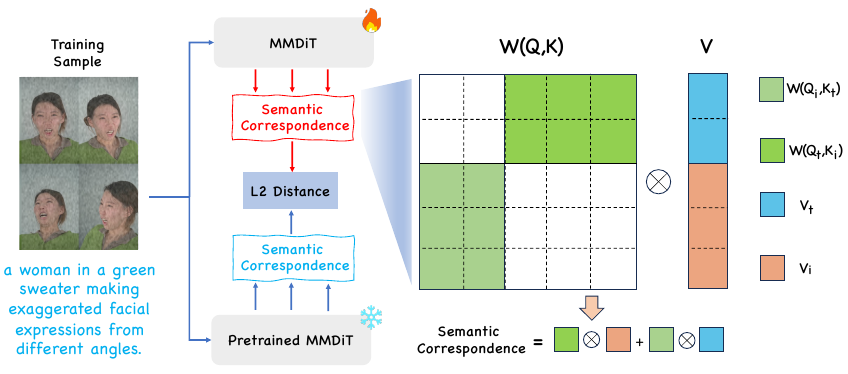}
    \vspace{-0.3cm}
    \caption{\textbf{Overview of Semantic Correspondence Alignment Loss.} Specifically, we minimize the L2 distance between semantic correspondences at each layer of the finetuned and pretrained MMDiT models for the same training sample, thereby explicitly constraining the finetuned model to retain the semantic control capabilities learned in pretraining.
    }
    \label{fig:SCA}
\end{figure*}

\subsection{Task Formulation}
Although existing human customization approaches have successfully demonstrated the ability to accurately preserve user identities and allow for various prompt-control, the generation of consistent multi-view customized images has yet to be thoroughly explored. In this work, we introduce a novel task: Multi-View Customization for human images. Given an image provided by the user and the prompt condition, our objective is to generate multiple customized images from different viewpoints that maintain high geometric and visual consistency across all perspectives, with robust identity similarity and prompt following.
\subsection{Priors-Guided In-Context Generation}
Previous works~\citep{lhhuang2024iclora, kang2025flux} have demonstrated the contextual learning capabilities of the DiT model, leveraging this to generate images with consistent appearances. However, they unexplore the potential to preserve the consistency in multi-view scenarios. It motivates us to investigate whether the model has a strong multi-view capability to foster more explicit multi-view consistency. To explore this, we propose utilizing in-context depth maps as priors to activate the multi-view consistency of the DiT model. 

Specifically, we adopt a two-stage approach as shown in \cref{fig:method}. In the first stage, our objective is to obtain multi-view depth maps that align with the provided prompt. Therefore, we leverage a pre-trained customization generator, like PuLID~\citep{guo2024pulid}, to perform initial sampling based on the user-provided image and prompt. Subsequently, we employ SMPL~\citep{goel2023humans} to fit a corresponding human body model and generate multi-view depth maps by rotating and rendering the fitted mesh. In the second stage, these multi-view human depth maps are arranged into a four-panel grid, which serves as an in-context conditioning signal for a pre-trained depth-conditioned model~\citep{flux1-dev}. This setup enables the synthesis of multi-view consistent customized images.

\subsection{Lightweight In-Context Learning}

With this in-context depth-conditioned framework, we can further fine-tune the depth-conditioned adapter with an in-context learning paradigm to enhance the geometry and visual consistency. Empirical results demonstrate that, only a small amount of training samples is required to activate the model's inherent capability for multi-view consistency. Specifically, we sample multi-view human depth-image pairs from the multi-view dataset like NeRSemble~\citep{kirschstein2023nersemble} and use the VLM model~\citep{bai2023qwenvlversatilevisionlanguagemodel} to generate captions that correspond to the respective prompts. These depth maps and images are then organized into a four-panel grid format, which is utilized for in-context learning during training.
To alleviate the adverse effects of background diversity, we further extract body masks $M$ from the depth maps and integrate them into the diffusion reconstruction loss as spatial priors,
\begin{equation}
    \begin{aligned}
     \mathcal{L}_{\text{diff}} = \mathbb{E}_{\x_t,t,\mathbf{c}, \epsilon \sim \mathcal{N} (\mathbf{0}, \mathbf{I})} [w_t||\epsilon - \epsilon_{\theta} (\x_t, t, \mathbf{c}) || \cdot M].
    \end{aligned}\label{eq:diff}
\end{equation}


\subsection{Semantic Correspondence Alignment}
While the multi-view consistency is significantly activated with the 3D priors, acquiring robust multi-view customization capability typically demands large-scale multi-dimensional datasets, making multi-view contextual learning under limited data prone to textual controllability degradation. To address this degradation of semantic capabilities, we propose a Semantic Correspondence Alignment Loss as shown in \cref{fig:SCA}.

Our primary goal is to preserve the model's original ability to respond to semantic textual inputs. To achieve this, we introduce a frozen branch of the pretrained model during training. Intuitively, aligning the correspondence between textual and visual tokens of the dual branch will facilitate the preservation of cross-modal semantic consistency without affecting its other behaviors. To the end, for the Query-Key-Value components in \cref{eq:attn_qkv} and \cref{eq:attn_eq}, we compute the semantic correspondence of $Q$, $K$ and $V$ in each layer $l$

\begin{equation}
    SC(Q^l,K^l,V^l) = A(Q^l_I, K^l_T, V^l_T) + A(Q^l_T, K^l_I, V^l_I).  
    \label{eq:d1}
\end{equation}
Then we minimize the L2 distance between all the semantic correspondence of the fine-tuned and pretrained MMDiT
\begin{equation}
    \mathcal{L}_{\text{SCA}} = \mathbb{E}_{\x_t,t,l}||SC^F - SC^P||_2,
    \label{eq:sca}
\end{equation}
where $SC^F$ and $SC^P$ are the semantic correspondence of the fine-tuned and pretrained MMDiT respectively. The alignment ensures that the model retains its semantic response capability while benefiting from multi-view consistency. Overall, our training loss is 
\begin{equation}
    \mathcal{L}_{\text{total}} = \mathcal{L}_{\text{diff}} + \mathcal{L}_{\text{SCA}}.
    \label{eq:loss}
\end{equation}

\begin{figure*}[ht]
    \centering
    \includegraphics[width=\linewidth]{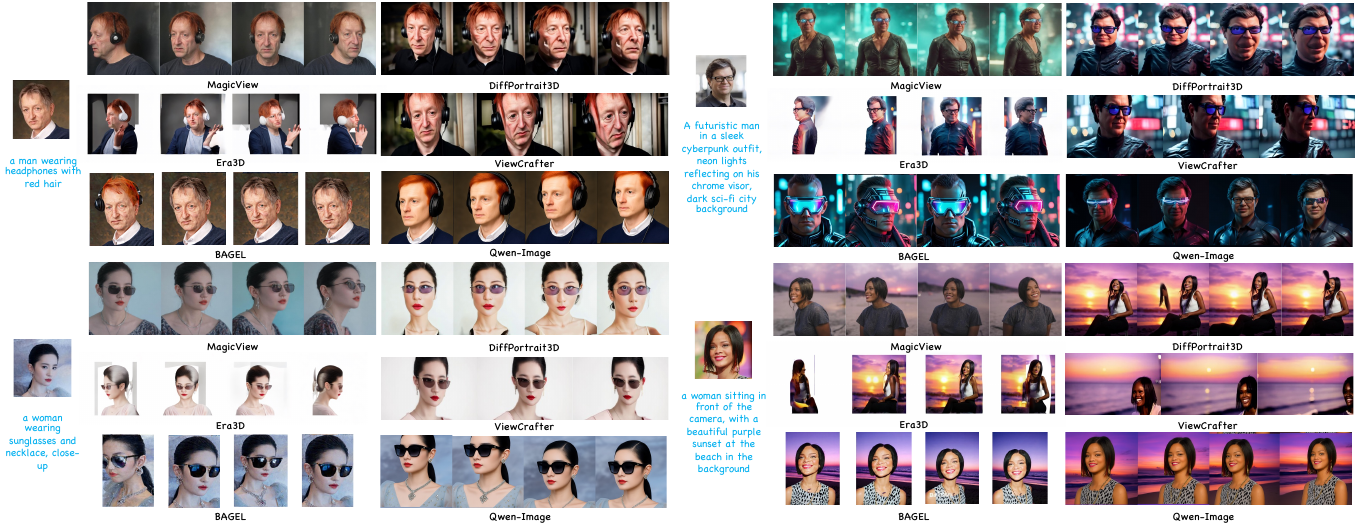}
        \vspace{-0.4cm}
    \caption{\textbf{Qualitative comparison.} DiffPortrait3D and Era3D exhibit limitations in maintaining geometric and visual consistency, especially with regard to full-body and background regions. Although ViewCrafter achieves improved scene modeling, it does so at the expense of geometric consistency in human representations. Besides, both BAGEL and Qwen-Image demonstrate suboptimal performance in terms of multi-view control. In contrast, our \modulespace achieves superior performance in both geometric fidelity and visual coherence across views. The baseline methods generate results using prompts that explicitly specify viewpoints, such as “left view”, “middle view”, and “right view”.    }
    \label{fig:exp}
\end{figure*}

\section{Experiments}

\begin{table}[t]
    \tablestyle{2pt}{1.15}
    \centering
\caption{\textbf{Quantitative comparison.} We conduct a comprehensive comparison including the ability to achieve high multi-view consistency, identity consistency with the reference image (\textit{i.e.}, ID Consistency and CLIP-I), and text alignment (\textit{i.e.}, CLIP-T).}
\label{tab:result}
    \begin{tabular}{ccccc} 
        \toprule
\textbf{Method} & \textbf{MV Cons.} ($\downarrow$) & \textbf{ID Cons.} ($\uparrow$) & \textbf{CLIP-T} ($\uparrow$) & \textbf{CLIP-I} ($\uparrow$)\\
\toprule
DiffPortrait3D~\citep{gu2024diffportrait3d} &7.887&0.7721&0.2588&0.6689 \\
Era3D~\citep{li2024era3d} & 6.383 & 0.6462 & 0.2421 & 0.6708 \\
ViewCrafter~\citep{yu2024viewcrafter} &6.945&0.5820&0.2358&0.7037 \\
BAGEL~\citep{deng2025emerging} &5.882&0.5623&0.2591&0.7289 \\
Qwen-image~\citep{wu2025qwen} &5.324&0.6433&0.2603&0.7422 \\ 
\midrule
\textbf{\module} &\textbf{3.697} &\textbf{0.7920} &  \textbf{0.2615} &\textbf{0.7793}  \\
        \bottomrule
    \end{tabular}
\end{table}

\begin{table}
    \caption{\textbf{User Study.} Our \modulespace achieves the best human preference compared with all baselines.}
    \tablestyle{1pt}{1}
    \centering
    \begin{tabular}{ccccc}
    \toprule
         Method & \textbf{MV Cons.}   & \textbf{Text Align.}  & \textbf{ID Cons.} & \textbf{Overall} \\
            \midrule
       Era3D & 10.33   &6.24&15.55&10.43 \\
     DiffPortrait3D &15.62&15.71&10.74&11.88\\
ViewCrafter&10.52&10.92&13.64 &14.23\\
        BAGEL&18.18&17.83&12.98 &18.32\\
        Qwen-Image&21.22&22.38& 17.42&20.58\\
        \midrule
    \textbf{\modulespace}&\textbf{31.80}&\textbf{26.92}&\textbf{29.67}&\textbf{24.56}\\
    \bottomrule
    \end{tabular}
\label{tab:user}
\end{table}

\subsection{Experimental Setup}
\textbf{Implementation details.}
We utilize the recently developed DiT model FLUX~\citep{flux1-dev} with a pre-trained personalized module PuLID~\citep{guo2024pulid} as our foundational model. 
Our training set comprises only 100 cases randomly sampled from NeRSemble~\citep{kirschstein2023nersemble}, a widely utilized multi-view human dataset. Please refer to the Appendix for more details.

\noindent
\textbf{Baselines.}
To the best of our knowledge, our approach is the first to support end-to-end customized multi-view human image generation. For comparison, we construct a two-stage baseline wherein PuLID is used for identity-specific image synthesis, followed by the novel view synthesis method based on image-to-3D reconstruction.
Specifically, we benchmark the proposed method against DiffPortrait3D~\citep{gu2024diffportrait3d} and Era3D~\citep{li2024era3d}, recent methods synthesizing 3D-consistent photo-realistic novel views. 
We also compare with ViewCrafter~\citep{yu2024viewcrafter}, a recent method synthesizing high-fidelity novel views of generic scenes. Besides, we compare with recent unified multi-modal model BAGEL~\citep{deng2025emerging} and Qwen-Image~\citep{wu2025qwen}.

\begin{figure*}[ht]
    \centering
    \includegraphics[width=.97\linewidth]{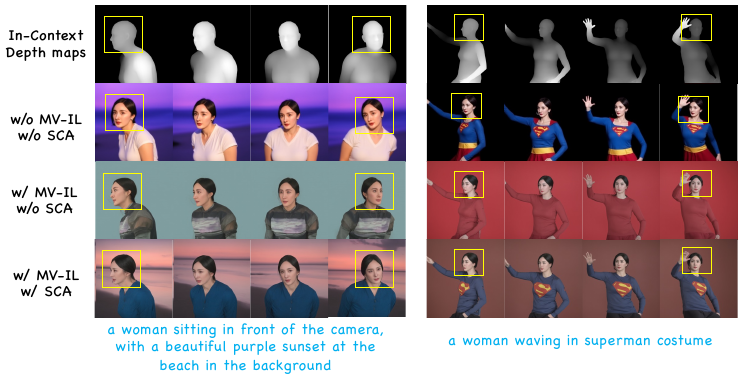}
    \vspace{-0.4cm}
    \caption{\textbf{Ablation study for multi-view in-context learning.}
    As shown, transitioning from the pretrained model (second row) to the multi-view in-context learning model (third row) significantly improves cross-view geometric consistency. The addition of the SCA module (fourth row) helps the finetuned model retain the pretrained model's ability to follow prompt-based semantic controls, such as identity, clothing, and background.
    }
    \vspace{-0.3cm}
    \label{fig:ablation}
\end{figure*}

\noindent
\textbf{Evaluation.}
We collect a diverse portrait test set from the internet which consists of 80 characters ensuring demographic representation, with 50 prompts for comprehensive motion evaluation.
To assess multi-view consistency, we follow the 3D consistency metric (MV Cons.) proposed in Pippo~\citep{kant2025pippo}. MV Cons. is calculated on a pair of images where landmarks are estimated first and then reprojected to each other for error calculation. More details are in Appendix.
The evaluation framework employs CLIP-T and CLIP-I for quantitative assessment of text and image alignment. To further measure identity preservation, we implement facial recognition embedding similarity metrics~\citep{an2021partial} complemented by specialized facial motion analysis protocols. 
\subsection{Main Results}
\noindent
\textbf{Qualitative Results.}
We present a qualitative assessment comparing \modulespace with baseline methods. 
As illustrated in \cref{fig:exp}, DiffPortrait3D struggles with maintaining full-body geometric consistency and coherent scene appearance, largely due to its design and training focus on portrait-level generation. Similarly, Era3D, by primarily targeting subject-centric multi-view synthesis, performs poorly in preserving visual consistency across scene elements. Although ViewCrafter achieves improved scene coherence, it does so at the cost of human body geometry, often leading to distorted multi-view results. For BAGEL and Qwen-Image, they both exhibit relatively poor performance in multi-view control and consistency. In contrast, our proposed method achieves strong geometric consistency for the entire human body, along with visually consistent scene modeling, including accessories and backgrounds, while simultaneously preserving identity fidelity and supporting prompt-guided customization. More results are included in the Appendix.


\noindent
\textbf{Quantitative Results.}
We report the results of the quantitative comparison in \cref{tab:result}. As observed, DiffPortrait3D exhibits limited multi-view consistency, highlighting its weakness in modeling coherent geometry across viewpoints. Although Era3D mitigates this issue to some extent, it still suffers from subpar performance in identity preservation and semantic control accuracy. Likewise, ViewCrafter shows inferior results in both identity consistency and CLIP-based semantic alignment. Although BAGEL and Qwen-Image improve multi-view consistency, the identity similarity has decreased. In contrast, our method consistently achieves superior performance across all key evaluation dimensions, including multi-view consistency, identity fidelity, and prompt-aligned semantic controllability.

\begin{table}
    \caption{\textbf{Quantitative comparison with ImagPose and MVDream.} 
    }
        \vspace{-0.2cm}
    \tablestyle{1.5pt}{1}
    \centering
    \begin{tabular}{c|ccc}
    \toprule
         Method & \textbf{MV Cons.} ($\downarrow$)  & \textbf{CLIP-T}  ($\uparrow$)& \textbf{ID Cons.} ($\uparrow$)\\
            \midrule
         ImagPose &4.882   &0.2385&0.7315\\
     MVDream&8.669   &0.2322&0.7286\\
        Ours &\textbf{3.697}  &\textbf{0.2604}&\textbf{0.7920}\\
    \bottomrule
    \end{tabular}
\label{tab:more_baseline}
\end{table}

\begin{table}
    \caption{\textbf{Quantitative ablation study of multi-view in-context learning.} 
    As observed in \cref{fig:ablation}, applying multi-view in-context learning improves multi-view consistency but degrades semantic controllability and identity consistency. Incorporating the SCA loss allows our model to recover these capabilities, balancing geometric consistency and semantic fidelity.
    }
    \tablestyle{3.5pt}{1.15}
    \centering
    \begin{tabular}{cc|ccc}
    \toprule
         MV-IL & SCA & \textbf{MV Cons.} ($\downarrow$)  & \textbf{CLIP-T}  ($\uparrow$)& \textbf{ID Cons.} ($\uparrow$)\\
            \midrule
        & &7.270   &0.2589&\textbf{0.7943}\\
    \checkmark& &3.928   &0.2271&0.7753\\
        \checkmark&\checkmark &\textbf{3.697}  &\textbf{0.2604}&0.7920\\
    \bottomrule
    \end{tabular}
\label{tab:ablation}
\end{table}

\begin{table*}[ht]
    \centering
        \caption{\textbf{Ablation study for the number of training iterations and samples.} $^*$ indicates the best trade-off setting.}
    \begin{subtable}{.5\linewidth}
    \tablestyle{3.5pt}{1.15}
    \centering
    \begin{tabular}{c|ccc}
    \toprule
         Num. & \textbf{MV Cons.} ($\downarrow$)  & \textbf{CLIP-T}  ($\uparrow$)& \textbf{ID Cons.} ($\uparrow$)\\
            \midrule
    200&5.553   &0.2596&0.7902\\
    400 &4.282 &0.2588&0.7899 \\
    800* &3.694 &\textbf{0.2599}&\textbf{0.7912}\\
    1600 &\textbf{3.685} &0.2579&0.7853 \\
    \bottomrule
    \end{tabular}
    \vspace{0.1cm}
    \caption{\textbf{Ablation study of training iterations.}
    }
\label{tab:iter}
\vspace{-0.4cm}
    \end{subtable}%
\begin{subtable}{.5\linewidth}
    \tablestyle{3.5pt}{1.15}
    \centering
    \begin{tabular}{c|ccc}
    \toprule
         Num. & \textbf{MV Cons.} ($\downarrow$)  & \textbf{CLIP-T}  ($\uparrow$)& \textbf{ID Cons.} ($\uparrow$)\\
            \midrule
    100* &3.697  &\textbf{0.2604}&\textbf{0.7920}\\
    200 &3.752 &0.2543&0.7899 \\
    400 &\textbf{3.688} &0.2522&0.7882 \\
    800 &3.694 &0.2599&0.7912\\
    \bottomrule
    \end{tabular}
    \vspace{0.1cm}
    \caption{\textbf{Ablation study of training samples.} 
    }
\label{tab:sample}
\vspace{-0.4cm}
    \end{subtable}%
\end{table*}

\subsection{Comparison with More Methods}
We provide comparisons with additional potential baselines, such as ImagPose~\citep{shen2024imagpose} and MVDream~\citep{shi2023mvdream}. Compared with ImagPose, our method enables the customization model to generate multi-view consistent images via lightweight adaptation with only 100 samples, while Imagpose requires ~85000 images with diverse viewpoints. Besides, Imagpose lacks the ability to control rich semantics, such as background changes via prompts, while our method preserves the precise semantic control as validated in \cref{tab:more_baseline}. For MVDream, it exhibits limited multiview consistency and semantic control, probably because it has difficulty generalizing to the human domain.

\noindent
\textbf{User Study.}
To further evaluate the effectiveness of our methodology, we conduct a human-centric assessment, comparing our approach with existing novel view synthesis methods. We recruit 25 evaluators to assess 40 sets of generated results. For each set, we present reference images alongside multi-view images produced using identical seeds across various methods. The quality of the generated multi-view images is evaluated based on four criteria: Multi-View Consistency, Text Alignment, ID Consistency, and Overall Quality.
As depicted in \cref{tab:user}, our \modulespace achieves higher user preference across all evaluative dimensions, underscoring its superior effectiveness.

\subsection{Ablation Study}
\textbf{Multi-View In-Context Learning.}
To validate the effectiveness of our proposed components, we conduct an ablation study presented in \cref{fig:ablation} and \cref{tab:ablation}. As shown in the figure, the pretrained model often generates head orientations inconsistent with the corresponding depth maps, leading to degraded geometric consistency across viewpoints. Incorporating multi-view in-context learning significantly alleviates this issue, resulting in more coherent human geometry across views. However, due to the limited diversity of scenes in the multi-view dataset, prompt-based semantic controllability, particularly for attributes such as background and clothing, tends to diminish. In contrast, Semantic Correspondence Alignment (SCA) loss effectively preserves fine- grained semantic control, ensuring that identity and prompt-relevant attributes are maintained throughout the generated views. More ablation studies are given in the Appendix.

\noindent
\textbf{Number of Training Iterations and Samples.}
To investigate the impact of training iterations on model performance, we present a quantitative comparison in \cref{tab:iter}. As indicated by the results, 800 iterations provide a favorable trade-off between efficiency and performance, and are therefore adopted in our final setting. Additionally, to validate the lightweight nature of our in-context learning framework, we evaluate the effect of varying the number of training samples. As shown in \cref{tab:sample}, our method achieves competitive performance with as few as 100 training samples, demonstrating its data efficiency and strong generalization capability under limited supervision.

\noindent
\textbf{Design of SCA.}
The core design of SCA is to preserve the original model’s semantic control (text-image correspondence) without compromising in-context learning during fine-tuning. In contrast, applying regularization on other aspects like full-attention features or model parameters leads to reduced multi-view consistency as shown in \cref{tab:sca}, since it disrupts in-context learning.

\begin{table}
    \caption{\textbf{Quantitative ablation study of different alignment loss.} 
    }
    \tablestyle{3.5pt}{1}
    \centering
    \begin{tabular}{c|ccc}
    \toprule
         Method & \textbf{MV Cons.} ($\downarrow$)  & \textbf{CLIP-T}  ($\uparrow$)& \textbf{ID Cons.} ($\uparrow$)\\
            \midrule
          Full-Attn.&5.433   &0.2593&0.7855\\
     Parameter&6.129   &0.2602&0.7678\\
        Ours &\textbf{3.697}  &\textbf{0.2604}&\textbf{0.7920}\\
    \bottomrule
    \end{tabular}
    \vspace{-0.2cm}
\label{tab:sca}
\end{table}
\section{Conclusion}
In this work, we introduce \module, a novel framework for multi-view human image customization from a single photograph. By investigating and leveraging in-context learning with multi-view depth priors, our method enhances DiT-based models' multi-view reasoning. We also propose a Semantic Correspondence Alignment Loss to preserve prompt controllability during fine-tuning. Extensive experiments show that \modulespace excels in identity fidelity and coherent multi-view synthesis, requiring minimal additional training. Our approach enables efficient, high-quality multi-view customization without large-scale multi-view datasets.

\section{Limitation}
While our approach demonstrates strong performance in generating multi-view consistent customized images, it also has certain limitations. In particular, it primarily relies on activating the in-context learning capabilities of the base model with multi-view priors, and therefore, its effectiveness is inherently constrained by the representational power of the underlying pre-trained model. Nevertheless, a key advantage of our framework lies in its modularity and transferability, making it easily adaptable to more powerful backbone models as they emerge, a direction we plan to explore in future work.

%
{
    \small
    \bibliographystyle{ieeenat_fullname}
    \bibliography{main}

@String{Computer = "{IEEE} Computer" }

@String{Springer = "Springer-Verlag" }

@misc{bai2023qwenvlversatilevisionlanguagemodel,
      title={Qwen-VL: A Versatile Vision-Language Model for Understanding, Localization, Text Reading, and Beyond}, 
      author={Jinze Bai and Shuai Bai and Shusheng Yang and Shijie Wang and Sinan Tan and Peng Wang and Junyang Lin and Chang Zhou and Jingren Zhou},
      year={2023},
      eprint={2308.12966},
      archivePrefix={arXiv},
      primaryClass={cs.CV},
      url={https://arxiv.org/abs/2308.12966}, 
}

@inproceedings{sohl2015deep,
  title={Deep unsupervised learning using nonequilibrium thermodynamics},
  author={Sohl-Dickstein, Jascha and Weiss, Eric and Maheswaranathan, Niru and Ganguli, Surya},
  booktitle={ICML},
  year={2015}
}

@article{podell2023sdxl,
  title={Sdxl: Improving latent diffusion models for high-resolution image synthesis},
  author={Podell, Dustin and English, Zion and Lacey, Kyle and Blattmann, Andreas and Dockhorn, Tim and M{\"u}ller, Jonas and Penna, Joe and Rombach, Robin},
  journal={arXiv preprint arXiv:2307.01952},
  year={2023}
}

@article{gal2022image,
  title={An image is worth one word: Personalizing text-to-image generation using textual inversion},
  author={Gal, Rinon and Alaluf, Yuval and Atzmon, Yuval and Patashnik, Or and Bermano, Amit H and Chechik, Gal and Cohen-Or, Daniel},
  journal={arXiv preprint arXiv:2208.01618},
  year={2022}
}

@article{ho2020denoising,
  title={Denoising diffusion probabilistic models},
  author={Ho, Jonathan and Jain, Ajay and Abbeel, Pieter},
  journal=NIPS,
  volume={33},
  pages={6840--6851},
  year={2020}
}

@article{hu2021lora,
  title={Lora: Low-rank adaptation of large language models},
  author={Hu, Edward J and Shen, Yelong and Wallis, Phillip and Allen-Zhu, Zeyuan and Li, Yuanzhi and Wang, Shean and Wang, Lu and Chen, Weizhu},
  journal={arXiv preprint arXiv:2106.09685},
  year={2021}
}

@article{wang2024instantid,
  title={Instantid: Zero-shot identity-preserving generation in seconds},
  author={Wang, Qixun and Bai, Xu and Wang, Haofan and Qin, Zekui and Chen, Anthony},
  journal={arXiv preprint arXiv:2401.07519},
  year={2024}
}

@inproceedings{li2024photomaker,
  title={Photomaker: Customizing realistic human photos via stacked id embedding},
  author={Li, Zhen and Cao, Mingdeng and Wang, Xintao and Qi, Zhongang and Cheng, Ming-Ming and Shan, Ying},
  booktitle={Proceedings of the IEEE/CVF Conference on Computer Vision and Pattern Recognition},
  pages={8640--8650},
  year={2024}
}

@article{guo2024pulid,
  title={PuLID: Pure and Lightning ID Customization via Contrastive Alignment},
  author={Guo, Zinan and Wu, Yanze and Chen, Zhuowei and Chen, Lang and He, Qian},
  journal={arXiv preprint arXiv:2404.16022},
  year={2024}
}

@inproceedings{an2021partial,
  title={Partial fc: Training 10 million identities on a single machine},
  author={An, Xiang and Zhu, Xuhan and Gao, Yuan and Xiao, Yang and Zhao, Yongle and Feng, Ziyong and Wu, Lan and Qin, Bin and Zhang, Ming and Zhang, Debing and others},
  booktitle={Proceedings of the IEEE/CVF International Conference on Computer Vision},
  pages={1445--1449},
  year={2021}
}

@article{gal2024lcm,
  title={Lcm-lookahead for encoder-based text-to-image personalization},
  author={Gal, Rinon and Lichter, Or and Richardson, Elad and Patashnik, Or and Bermano, Amit H and Chechik, Gal and Cohen-Or, Daniel},
  journal={arXiv preprint arXiv:2404.03620},
  year={2024}
}

@inproceedings{peng2024portraitbooth,
  title={Portraitbooth: A versatile portrait model for fast identity-preserved personalization},
  author={Peng, Xu and Zhu, Junwei and Jiang, Boyuan and Tai, Ying and Luo, Donghao and Zhang, Jiangning and Lin, Wei and Jin, Taisong and Wang, Chengjie and Ji, Rongrong},
  booktitle={Proceedings of the IEEE/CVF Conference on Computer Vision and Pattern Recognition},
  pages={27080--27090},
  year={2024}
}

@inproceedings{ma2024subject,
  title={Subject-diffusion: Open domain personalized text-to-image generation without test-time fine-tuning},
  author={Ma, Jian and Liang, Junhao and Chen, Chen and Lu, Haonan},
  booktitle={ACM SIGGRAPH 2024 Conference Papers},
  pages={1--12},
  year={2024}
}

@inproceedings{valevski2023face0,
  title={Face0: Instantaneously conditioning a text-to-image model on a face},
  author={Valevski, Dani and Lumen, Danny and Matias, Yossi and Leviathan, Yaniv},
  booktitle={SIGGRAPH Asia 2023 Conference Papers},
  pages={1--10},
  year={2023}
}

@article{xiao2024fastcomposer,
  title={Fastcomposer: Tuning-free multi-subject image generation with localized attention},
  author={Xiao, Guangxuan and Yin, Tianwei and Freeman, William T and Durand, Fr{\'e}do and Han, Song},
  journal={International Journal of Computer Vision},
  pages={1--20},
  year={2024},
  publisher={Springer}
}

@inproceedings{li2023few,
  title={Few-shot Hybrid Domain Adaptation of Image Generator},
  author={Li, Hengjia and Liu, Yang and Xia, Linxuan and Lin, Yuqi and Wang, Wenxiao and Zheng, Tu and Yang, Zheng and Zhong, Xiaohui and Ren, Xiaobo and He, Xiaofei},
  booktitle={The Twelfth International Conference on Learning Representations},
  year={2023}
}

@article{li2024unihda,
  title={UniHDA: Towards Universal Hybrid Domain Adaptation of Image Generators},
  author={Li, Hengjia and Liu, Yang and Lin, Yuqi and Zhang, Zhanwei and Zhao, Yibo and Zheng, Tu and Yang, Zheng and Jiang, Yuchun and Wu, Boxi and Cai, Deng and others},
  journal={arXiv preprint arXiv:2401.12596},
  year={2024}
}

@article{li2024personalvideo,
  title={PersonalVideo: High ID-Fidelity Video Customization without Dynamic and Semantic Degradation},
  author={Li, Hengjia and Qiu, Haonan and Zhang, Shiwei and Wang, Xiang and Wei, Yujie and Li, Zekun and Zhang, Yingya and Wu, Boxi and Cai, Deng},
  journal={arXiv preprint arXiv:2411.17048},
  year={2024}
}

@article{ipadapter,
  title={Ip-adapter: Text compatible image prompt adapter for text-to-image diffusion models},
  author={Ye, Hu and Zhang, Jun and Liu, Sibo and Han, Xiao and Yang, Wei},
  journal={arXiv preprint arXiv:2308.06721},
  year={2023}
}

@inproceedings{dit,
  title={Scalable diffusion models with transformers},
  author={Peebles, William and Xie, Saining},
  booktitle={Proceedings of the IEEE/CVF International Conference on Computer Vision},
  pages={4195--4205},
  year={2023}
}

@misc{flux1-dev,
  author       = {Black Forest Labs},
  title        = {FLUX.1-dev},
  howpublished = {\url{https://huggingface.co/black-forest-labs/FLUX.1-dev}},
  year         = {2024}
}

@article{kant2025pippo,
  title={Pippo: High-Resolution Multi-View Humans from a Single Image},
  author={Kant, Yash and Weber, Ethan and Kim, Jin Kyu and Khirodkar, Rawal and Zhaoen, Su and Martinez, Julieta and Gilitschenski, Igor and Saito, Shunsuke and Bagautdinov, Timur},
  journal={arXiv preprint arXiv:2502.07785},
  year={2025}
}

@article{yu2024viewcrafter,
  title={Viewcrafter: Taming video diffusion models for high-fidelity novel view synthesis},
  author={Yu, Wangbo and Xing, Jinbo and Yuan, Li and Hu, Wenbo and Li, Xiaoyu and Huang, Zhipeng and Gao, Xiangjun and Wong, Tien-Tsin and Shan, Ying and Tian, Yonghong},
  journal={arXiv preprint arXiv:2409.02048},
  year={2024}
}

@inproceedings{gu2024diffportrait3d,
  title={Diffportrait3d: Controllable diffusion for zero-shot portrait view synthesis},
  author={Gu, Yuming and Xu, Hongyi and Xie, You and Song, Guoxian and Shi, Yichun and Chang, Di and Yang, Jing and Luo, Linjie},
  booktitle={Proceedings of the IEEE/CVF Conference on Computer Vision and Pattern Recognition},
  pages={10456--10465},
  year={2024}
}

@article{li2024era3d,
  title={Era3d: high-resolution multiview diffusion using efficient row-wise attention},
  author={Li, Peng and Liu, Yuan and Long, Xiaoxiao and Zhang, Feihu and Lin, Cheng and Li, Mengfei and Qi, Xingqun and Zhang, Shanghang and Xue, Wei and Luo, Wenhan and others},
  journal={Advances in Neural Information Processing Systems},
  volume={37},
  pages={55975--56000},
  year={2024}
}

@article{kirschstein2023nersemble,
  title={Nersemble: Multi-view radiance field reconstruction of human heads},
  author={Kirschstein, Tobias and Qian, Shenhan and Giebenhain, Simon and Walter, Tim and Nie{\ss}ner, Matthias},
  journal={ACM Transactions on Graphics (TOG)},
  volume={42},
  number={4},
  pages={1--14},
  year={2023},
  publisher={ACM New York, NY, USA}
}

@article{long2023wonder3d,
  title={Wonder3D: Single Image to 3D using Cross-Domain Diffusion},
  author={Long, Xiaoxiao and Guo, Yuan-Chen and Lin, Cheng and Liu, Yuan and Dou, Zhiyang and Liu, Lingjie and Ma, Yuexin and Zhang, Song-Hai and Habermann, Marc and Theobalt, Christian and others},
  journal={arXiv preprint arXiv:2310.15008},
  year={2023}
}

@article{shi2023mvdream,
  title={Mvdream: Multi-view diffusion for 3d generation},
  author={Shi, Yichun and Wang, Peng and Ye, Jianglong and Long, Mai and Li, Kejie and Yang, Xiao},
  journal={arXiv preprint arXiv:2308.16512},
  year={2023}
}

@article{shi2023zero123++,
  title={Zero123++: a single image to consistent multi-view diffusion base model},
  author={Shi, Ruoxi and Chen, Hansheng and Zhang, Zhuoyang and Liu, Minghua and Xu, Chao and Wei, Xinyue and Chen, Linghao and Zeng, Chong and Su, Hao},
  journal={arXiv preprint arXiv:2310.15110},
  year={2023}
}

@article{tang2023MVDiffusion,
  title={MVDiffusion: Enabling Holistic Multi-view Image Generation with Correspondence-Aware Diffusion},
  author={Tang, Shitao and Zhang, Fuayng and Chen, Jiacheng and Wang, Peng and Yasutaka, Furukawa},
  journal={arXiv preprint 2307.01097},
  year={2023}
}

@article{voleti2024sv3d,
  title={Sv3d: Novel multi-view synthesis and 3d generation from a single image using latent video diffusion},
  author={Voleti, Vikram and Yao, Chun-Han and Boss, Mark and Letts, Adam and Pankratz, David and Tochilkin, Dmitry and Laforte, Christian and Rombach, Robin and Jampani, Varun},
  journal={arXiv preprint arXiv:2403.12008},
  year={2024}
}

@article{chen2024v3d,
  title={V3D: Video Diffusion Models are Effective 3D Generators},
  author={Chen, Zilong and Wang, Yikai and Wang, Feng and Wang, Zhengyi and Liu, Huaping},
  journal={arXiv preprint arXiv:2403.06738},
  year={2024}
}

@inproceedings{tseng2023consistent,
  title={Consistent view synthesis with pose-guided diffusion models},
  author={Tseng, Hung-Yu and Li, Qinbo and Kim, Changil and Alsisan, Suhib and Huang, Jia-Bin and Kopf, Johannes},
  booktitle={Proceedings of the IEEE/CVF Conference on Computer Vision and Pattern Recognition},
  pages={16773--16783},
  year={2023}
}

@article{kant2024spad,
  title={SPAD: Spatially Aware Multiview Diffusers},
  author={Kant, Yash and Wu, Ziyi and Vasilkovsky, Michael and Qian, Guocheng and Ren, Jian and Guler, Riza Alp and Ghanem, Bernard and Tulyakov, Sergey and Gilitschenski, Igor and Siarohin, Aliaksandr},
  journal={arXiv preprint arXiv:2402.05235},
  year={2024}
}

@article{kang2025flux,
  title={Flux Already Knows-Activating Subject-Driven Image Generation without Training},
  author={Kang, Hao and Fotiadis, Stathi and Jiang, Liming and Yan, Qing and Jia, Yumin and Liu, Zichuan and Chong, Min Jin and Lu, Xin},
  journal={arXiv preprint arXiv:2504.11478},
  year={2025}
}

@inproceedings{esser2024scaling,
  title={Scaling rectified flow transformers for high-resolution image synthesis},
  author={Esser, Patrick and Kulal, Sumith and Blattmann, Andreas and Entezari, Rahim and M{\"u}ller, Jonas and Saini, Harry and Levi, Yam and Lorenz, Dominik and Sauer, Axel and Boesel, Frederic and others},
  booktitle={Forty-first International Conference on Machine Learning},
  year={2024}
}

@misc{black2024flux,
  author = {Black Forest Labs},
  title = {FLUX.1-dev},
  year = {2024},
  howpublished = {https://huggingface.co/black-forest-labs/FLUX.1-dev},
  note = {Accessed: 2025-01-30}
}

@misc{openai2023dalle3,
  title = {DALL·E 3},
  author = {OpenAI},
  year = {2023},
  howpublished = {\url{https://openai.com/dall-e-3}},
  note = {Accessed: 2025-01-30}
}

@article{lhhuang2024iclora,
  title={In-Context LoRA for Diffusion Transformers},
  author={Huang, Lianghua and Wang, Wei and Wu, Zhi-Fan and Shi, Yupeng and Dou, Huanzhang and Liang, Chen and Feng, Yutong and Liu, Yu and Zhou, Jingren},
  journal={arXiv preprint arxiv:2410.23775},
  year={2024}
}

@inproceedings{kumari2024customizing,
  title={Customizing Text-to-Image Diffusion with Object Viewpoint Control},
  author={Kumari, Nupur and Su, Grace and Zhang, Richard and Park, Taesung and Shechtman, Eli and Zhu, Jun-Yan},
  booktitle={SIGGRAPH Asia 2024 Conference Papers},
  pages={1--13},
  year={2024}
}

@inproceedings{goel2023humans,
  title={Humans in 4D: Reconstructing and tracking humans with transformers},
  author={Goel, Shubham and Pavlakos, Georgios and Rajasegaran, Jathushan and Kanazawa, Angjoo and Malik, Jitendra},
  booktitle={Proceedings of the IEEE/CVF International Conference on Computer Vision},
  pages={14783--14794},
  year={2023}
}

@article{li2025magicid,
  title={MagicID: Hybrid Preference Optimization for ID-Consistent and Dynamic-Preserved Video Customization},
  author={Li, Hengjia and Jiang, Lifan and Xiao, Xi and Wang, Tianyang and Yi, Hongwei and Wu, Boxi and Cai, Deng},
  journal={arXiv preprint arXiv:2503.12689},
  year={2025}
}

@article{yang2024lora,
  title={Lora-composer: Leveraging low-rank adaptation for multi-concept customization in training-free diffusion models},
  author={Yang, Yang and Wang, Wen and Peng, Liang and Song, Chaotian and Chen, Yao and Li, Hengjia and Yang, Xiaolong and Lu, Qinglin and Cai, Deng and Wu, Boxi and others},
  journal={arXiv preprint arXiv:2403.11627},
  year={2024}
}

@article{li2024gca,
  title={GCA-3D: Towards Generalized and Consistent Domain Adaptation of 3D Generators},
  author={Li, Hengjia and Liu, Yang and Zhao, Yibo and Cheng, Haoran and Yang, Yang and Xia, Linxuan and Luo, Zekai and Qiu, Qibo and Wu, Boxi and Zheng, Tu and others},
  journal={arXiv preprint arXiv:2412.15491},
  year={2024}
}

@article{deng2025emerging,
  title={Emerging properties in unified multimodal pretraining},
  author={Deng, Chaorui and Zhu, Deyao and Li, Kunchang and Gou, Chenhui and Li, Feng and Wang, Zeyu and Zhong, Shu and Yu, Weihao and Nie, Xiaonan and Song, Ziang and others},
  journal={arXiv preprint arXiv:2505.14683},
  year={2025}
}

@article{wu2025qwen,
  title={Qwen-image technical report},
  author={Wu, Chenfei and Li, Jiahao and Zhou, Jingren and Lin, Junyang and Gao, Kaiyuan and Yan, Kun and Yin, Sheng-ming and Bai, Shuai and Xu, Xiao and Chen, Yilei and others},
  journal={arXiv preprint arXiv:2508.02324},
  year={2025}
}

@article{shen2024imagpose,
  title={Imagpose: A unified conditional framework for pose-guided person generation},
  author={Shen, Fei and Tang, Jinhui},
  journal={Advances in neural information processing systems},
  volume={37},
  pages={6246--6266},
  year={2024}
}

@article{chen2023photoverse,
  title={Photoverse: Tuning-free image customization with text-to-image diffusion models},
  author={Chen, Li and Zhao, Mengyi and Liu, Yiheng and Ding, Mingxu and Song, Yangyang and Wang, Shizun and Wang, Xu and Yang, Hao and Liu, Jing and Du, Kang and others},
  journal={arXiv preprint arXiv:2309.05793},
  year={2023}
}

@article{xu2025scalar,
  title={Scalar: Scale-wise controllable visual autoregressive learning},
  author={Xu, Ryan and Jin, Dongyang and Bai, Yancheng and Lan, Rui and Duan, Xu and Sun, Lei and Chu, Xiangxiang},
  year={2025}
}

@misc{jin2025scar,
      title={Semantic Context Matters: Improving Conditioning for Autoregressive Models}, 
      author={Jin, Dongyang and Xu, Ryan and Zeng, Jianhao and Lan, Rui and Bai, Yancheng and Sun, Lei and Chu, Xiangxiang},
      year={2025},
      journal={arXiv preprint arXiv:2511.14063},
}

@article{lan2025flux,
  title={Flux-text: A simple and advanced diffusion transformer baseline for scene text editing},
  author={Lan, Rui and Bai, Yancheng and Duan, Xu and Li, Mingxing and Jin, Dongyang and Xu, Ryan and Sun, Lei and Chu, Xiangxiang},
  journal={arXiv preprint arXiv:2505.03329},
  year={2025}
}
}

\clearpage
\setcounter{page}{1}
\maketitlesupplementary

\section{Implementation Details}
For the NeRSemble dataset, we employ Qwen-VL~\citep{bai2023qwenvlversatilevisionlanguagemodel} to generate descriptive captions.
During training, we learn the LoRA for 800 iterations with a learning rate of 1e-4 with batch size 1. We employ the AdamW optimizer with a weight decay parameter of 1e-2. 
The epsilon is set to the default 1e-8 and the weight decay is set to 1e-2. During inference, we use 50 steps of DDIM sampler and classifier-free guidance with a scale of 7.5. We generate multi-view images with 512 × 512 spatial resolution. We used Lora rank 128, following the common setting of FLUX finetuning. All experiments are conducted on a single NVIDIA A800 GPU. Our code will be open-source.
\section{Evaluation Metric for Multi-View Consistency}
To evaluate the multi-view consistency of the generated results, we adopt the re-projection error metric proposed in Pippo~\citep{kant2025pippo}. Specifically, we first estimate facial landmarks from the generated images and then establish pairwise correspondences of these landmarks across different views. Based on these correspondences, we apply Triangulation using the Direct Linear Transformation algorithm to recover the 3D positions of each landmark. Finally, we reproject the 3D landmarks back onto each view and compute the Reprojection Error as the L2 distance between the original 2D landmark and the reprojected point, normalized by the image resolution. The final RE score is obtained by averaging this error across all views and landmarks.

\section{Discussion of SMPL Fitting}
\begin{figure*}[t]
    \centering
    \includegraphics[width=\linewidth]{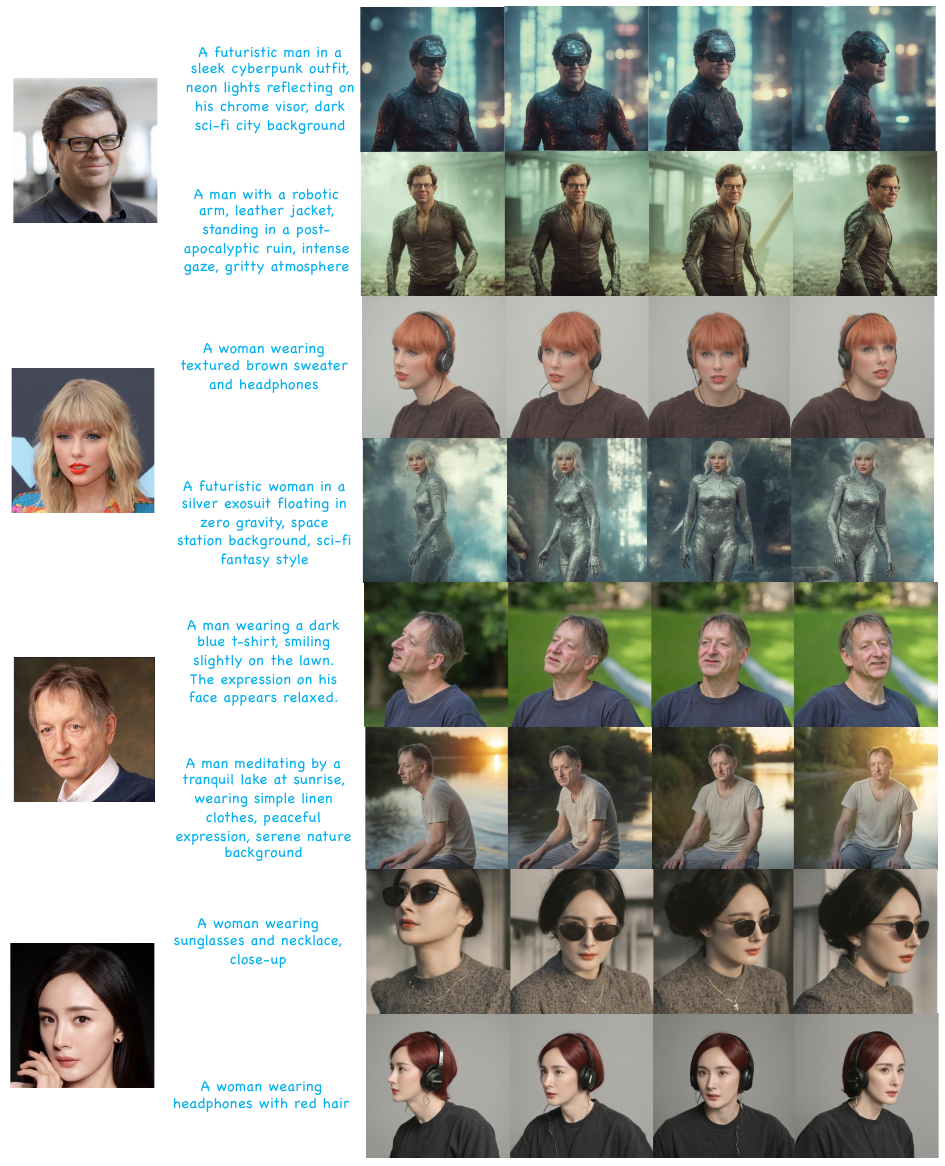}
    \vspace{-0.3cm}
    \caption{\textbf{More results of MagicView.}
    }
    \vspace{-0.6cm}
    \label{fig:add}
\end{figure*}

SMPL may introduce errors during body fitting, for instance, the body shape may undergo slight changes, and features like hair may not be accurately modeled. However, this does not affect the multi-view customization in the second stage, since our goal is to provide an initial depth condition to ensure robust multi-view consistency and reasonable semantic alignment and it is not necessary for the body shape in Step 2 to precisely match the single-view image in Step 1. For example, we can control features like hair through prompts, as shown in \cref{fig:exp} and more results in \cref{fig:add}. In practice, our method demonstrates excellent multi-view consistency and prompt adherence. On the other hand, it does not introduce excessive inference costs, with a total GPU usage of 9.8GB and a runtime of 0.45s.

\section{Difference from IC-LoRA.}
In practice, IC-LoRA~\citep{lhhuang2024iclora} boils down to fine-tuning the DiT on concatenated images, which is a simple yet widely-applicable technique. However, to our knowledge, there is no multi-view conditional generation model built on IC-LoRA at the time of submission. Our work provides solid evidence of the success of in-context learning in transferring multi-view generation to a generic model. On the other hand, our experiments demonstrate the insufficiency of using in-context learning alone, which has distinct contributions not covered by IC-LoRA.

\section{Ethics Statement}
Our main objective in this work is to empower novice users to generate visual content creatively and flexibly. However, we acknowledge the potential for misuse in creating fake or harmful content with our technology. Therefore, we believe it's essential to develop and implement tools to detect biases and malicious use cases to promote safe and equitable usage.

\section{Reproducibility Statement}
We make the following efforts to ensure the reproducibility of MagicView: (1) Our training and inference codes together with the trained model weights will be publicly available. (2) We provide training details in the appendix, which is easy to follow. (3) We provide the details of the human evaluation setups.

\section{LLM Usage Statement}
Large Language Models (LLMs), specifically OpenAI’s GPT-5, were employed as a general-purpose assistive tool during the preparation of this paper. The model was primarily used for:
(1) Language polishing – refining grammar, improving clarity, and adjusting tone to meet academic writing standards.
(2) Formatting support – generating LaTeX table templates, figure captions, and consistent section structuring.
All core research activities—including problem formulation, theoretical development, model design, experiments, analysis, and conclusions—were entirely conceived and executed by the authors. The LLM was not used for generating original research ideas, deriving results, or writing substantive scientific content.

\end{document}